\pdfoutput=1
\documentclass[11pt]{article}

\usepackage[final]{acl}

\usepackage{times}
\usepackage{latexsym}
\usepackage{booktabs}
\usepackage{amsmath}
\usepackage{tikz}

\usepackage[dvipsnames]{xcolor}
\usepackage[T1]{fontenc}
\usepackage[utf8]{inputenc}
\usepackage{microtype}

\usepackage{inconsolata}
\usepackage{listings}
\usepackage{graphicx}
\usepackage{float}

\usepackage{fancyhdr}
\usepackage{graphicx}
\usepackage{eso-pic}
\usepackage{threeparttable}
\usepackage{multirow}
\usepackage{setspace}
\usepackage{enumitem}      % adjust spacing in enums
\usepackage{graphicx}
\usepackage{epstopdf}
\usepackage{float}
\usepackage{amsmath,amsthm}
\usepackage{float}
\usepackage{booktabs}
\usepackage{multirow}
\usepackage{rotating}
\usepackage{pbox}
\usepackage{adjustbox}
\usepackage{tabularx}
\usepackage{calc}
\usepackage{setspace}
\usepackage{textcomp}
\setlist{nolistsep,leftmargin=*}
\title{Do Slides Help? Multi-modal Context for
Automatic Transcription of Conference Talks}

\author{\textbf{Supriti Sinhamahapatra} \hspace{2cm} \textbf{Jan Niehues} \\
         Karlsruhe Institute of Technology, Germany\\   \texttt{\{supriti.sinhamahapatra, jan.niehues\}@kit.edu}}

\begin{document}
\maketitle 
\begin{abstract}

State-of-the-art (SOTA) Automatic Speech Recognition (ASR) systems primarily rely on acoustic information while disregarding additional multi-modal context. However, visual information are essential in disambiguation and adaptation. 
While most work focus on speaker images to handle noise conditions, this work also focuses on integrating presentation slides for the use cases of scientific presentation.

In a first step, we create a  benchmark for multi-modal presentation including an automatic analysis of transcribing domain-specific terminology. Next, we explore methods for augmenting speech models with multi-modal information. 
We mitigate the lack of datasets with accompanying slides  by a suitable approach of data augmentation.
Finally, we train a model using the augmented dataset, resulting in  a relative reduction in word error rate of approximately 34\%, across all words and 35\%, for domain-specific terms compared to the baseline model. Our implementation  is available \footnote{http://doi.org/10.6084/m9.figshare.30158932}.
    %  We evaluate our approach on conference talks which are often accompanied by slides, and contain domain-specific words, potentially useful for improved ASR transcription.
    % We augment existing data set with additional information obtained from conference video recordings and show that the integration of this information to the models, improve the quality of transcription. For Whisper, SALMONN when integrated with information from slides, the word error rate on the domain-specific words reduces upto 48\%, 55\% respectively. 
\end{abstract}

\section{Introduction}
\label{sec:intro}

% JAN: My suggest copied from your part

\begin{figure}[t]
  \centering
  \includegraphics[scale= 0.48]{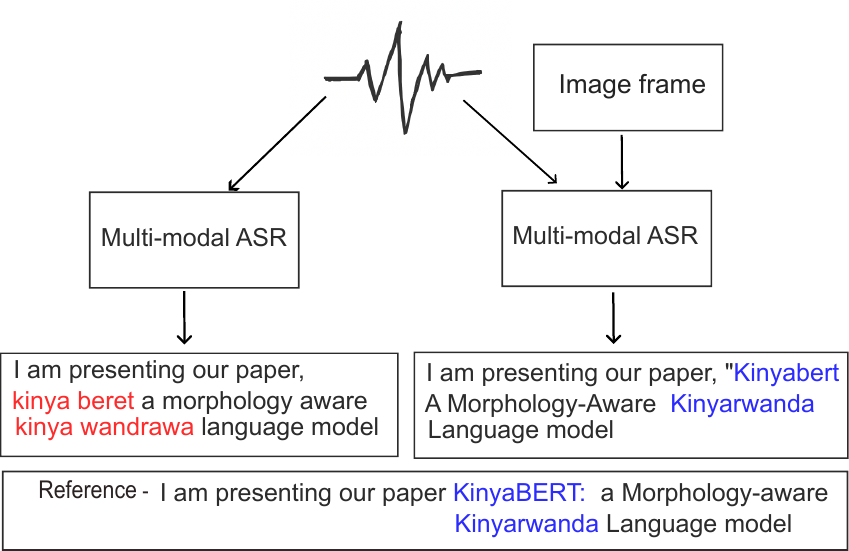}
  \caption{An example of ASR transcription before and after using multi-modal input. Left: ASR baseline makes mistakes (in red) for multiple words. Right: Model correctly transcribes words (in blue) using multi-modal inputs. }
  \label{fig:example}
\end{figure}

Automatic Speech Recognition (ASR) like many other NLP tasks are currently solved by using pre-trained models rather than learning models from scratch \cite{han2021pre}. 
Although modern ASR systems have an overall similar to human performance on general data yet one important challenge remains in accurately transcribing specialized vocabulary for example, in academic settings. Figure \ref{fig:example} illustrates a challenge for current ASR systems. A system  relying on only audio 
% (i.e. SALMONN \cite{tang2023salmonn}) 
is not able to correctly transcribe the domain-specific terms Kenya-Birth and Kenya Rwandan (in red).

As conference talks and lectures often include presentation slides, humans can correctly identify these words by using this additional context. Therefore, we propose to integrate visual context (slides) into existing state-of-the-art ASR system and enable them to also exploit this context. As shown on the right side of Figure \ref{fig:example}, the final model is able to properly transcribe  these words as  Kinyabert and Kinyarwanda (in blue)
when the correct words are presented to the model in the additional information provided from the accompanying slides of the talk. 

In a first step, we extend an existing benchmark, the ACL dataset \cite{salesky-etal-2023-evaluating} with additional slide context, as well as a, target automatic evaluation for domain-specific terms to evaluate this assumption. Furthermore, we verify our assumption that these terms are challenging for SOTA models like Whisper \cite{radford2023robust}, Phi-4-multimodal \cite{abouelenin2025phi} and SALMONN \cite{tang2023salmonn}.

When integrating visual context into ASR models to handle domain-specific words, we want to keep the strong SOTA performance of current large-scale models. Therefore, we focus on approach that can add this ability to existing models. One interesting aspect of current models is their ability to handle zero-shot task. To this end, we first propose a zero-shot integration that is already able to exploit the visual context. 

In a second step, we investigate methods to train the model to better integrate the contextual information. This gives rise to the challenge that we need dedicated training data for this scenario. We address this problem by using large language models (LLMs) to augment ASR training data with presentation slides.

%Motivated by this, we evaluate pre-trained models capabilities on transcribing domain-specific words and integrate the model with additional multi-modal information from slides, which are used as reference by presenters.

% This task is particularly challenging since we need to figure out more on what information would be helpful for Whisper and how to 
% This paper addresses two challenges in our approach.
% \begin{itemize}
    % \item Information Extraction: Since we are motivated to induce scientific information from the accompanying slides used by the authors as reference while presenting their work, we need to find strategies to extract relevant information from the slides. The challenge is, along with the relevant information, the slides also consist of general information. We show that with our approach of information extraction and filtration Whisper achieves an lower word error rates (WER) on the scientific terms.  
    % \item Model Evaluation: The second challenge is to find a proper way to evaluate Whisper.  Traditionally WER is used to evaluate ASR model performances by giving equal importance to every word present in the transcript. Since our interest lies specifically on Whisper performances on the special word, we find WER while only considering special words. 
    
The primary contributions of this paper are:
\begin{itemize}
    \item  Analysing the ability of ASR to transcribe domain-specific words, particularly from scientific talks.
    % In particular, we investigate model transcriptions on domain-specific words from scientific talks. 
    % We use the terms special words and domain-specific words interchangeably in this paper.
    \item   Integration of multi-modal information into existing pre-trained models. %Domain-specific relevant information is generated or extracted from conference video recordings and augmented to a model to improve transcription quality.

    \item Application of training approaches  with augmented data to improve the transcription on domain-specific terms.
    % Compare  with an approach where we simply find the special words from the manual transcriptions and as them as prompts.

\end{itemize}

 % This paper is structured as follows. First, we include the related work in Section \ref{sec:relwork} followed by description of the approaches of data integration in Section \ref{sec:approach}. Next, we provide a background of the model used and context integration to the model in Section \ref{sec:model} and  Finally, we include the experimental setup in Section \ref{sec:Experiment}, our evaluation strategies and our results in Sections \ref{sec:metric} and \ref{sec:result} respectively.

 \section{Related Work}\label{sec:relwork}
% {\color{blue}include comparisons with the following:
% Mala-asr: Multimedia-assisted llm-based asr
% SlideSpeech: A Large Scale Slide-Enriched Audio-Visual Corpus
% a. Tree-constrained pointer generator with graph neural network encodings for contextual speech recognition
% M3AV: A Multimodal, Multi genre, and Multipurpose Audio-Visual Academic Lecture Dataset". ACL 2023}

% Resturcture in according to contribution - Slide related ASR - Data augmentation with images 

% Researchers have always been interested in methods of using SOTA models in various tasks, either to improve their performance or to efficiently integrate them to some other task. 
 There are multiple work where model performance is improved by additional information integration.
  In this regard, \cite{maergner2012unsupervised} create a lecture specific vocabulary, based on the content of the related documents of the lectures. Construction of a vocabulary with relevant content improves the model performance and results in a reduced word error rate of up to 25 percent. 

Additionally, combining modalities for the improvement of ASR is also considered in the literature. Starting from Hidden Markov model for speech recognition and manually created features represented visual components, combining modalities were also considered for the task of establishing relation between words and non-linguistic context \cite{fleischman2008grounded} to compensate data deficiency. 
Later extraction of visual feature from videos using deep learning architectures was incorporated into ASR models on open-domain videos \cite{miao2016open}. These approaches are extended with  SOTA sequence to sequence model \cite{gupta2017visual} which helped to extract relevant context information from the videos for ASR.
  \cite{sun2022tree}  proposes using words from slides and presents GNN encoding using tree-RNN  for contextual speech recognition. In addition, \cite{huber2025continuously} performs a technique of continuous learning of new words in ASR from slides.

% In recent times,
% with the development of the end-to-end architectures, 
% transformer encoder decoder models are used to perform visual context-aware ASR. In the paper \cite{ghorbani2021listen} the authors use two separate encoders for encoding audio and visual modalities and a single decoder to handle the information from the encoders to generate improved transcription. 
% Other work include correcting ASR transcriptions with the help of visual modality \cite{srinivasan2020looking} where it is shown that, just by masking acoustic signals in the audio and providing relevant information from the visual modality is sufficient to make the model generate better transcription. For this approach, the authors uses a sequence-to-sequence encoder decoder model with a early decoder fusion strategy, where the visual features and the audio input vector from the encoder hidden state is concatenated and passed to the decoder.
Automatic speech recognition has made a significant progress in recent years by generating accurate transcriptions. Whisper~\cite{radford2023robust} has made it possible to generate better transcriptions on unseen datasets.
However, transcribing domain-specific datasets or low resource datasets, abbreviations, disfluencies still posses challenge for the SOTA ASR models\cite{ma2023adapting}. 
Many approaches focus on fusing audio and visual modalities to address challenges such as proper name transcription, error correction, noisy environments, and multi-modal context \cite{peng2023prompting},\cite{kumar2023visual}. 

In recent work, the integration of presentation slides into Multi-modal ASR has gained attention due to the potential benefits of leveraging visual information to improve transcription.  
The SLIDESPEECH dataset \cite{wang2024slidespeech}  a large scale audio-visual corpus enriched with slides is  created from online conference videos. However only a  part of their dataset is transcribed and synchronized with the slides. In a previous work, \cite{yang2024mala} creates a multi-modal-assisted LLM-based ASR model, and uses SLIDESPEECH dataset along with its accessible keywords provided with the dataset to enhance the ASR performance. In contrast to this paper, we explore a strategy to augment existing domain-specific speech-only datasets with images of slides, to enhance model performance on domain-specific vocabulary. Unlike \cite{yang2024mala}, we further demonstrate that incorporating images rather than textual context yields additional improvements in ASR performance.
Similar to the SLIDESPEECH dataset \cite{wang2024slideavsr} creates a dataset SlideAVSR, using scientific paper explanation videos.   They propose a  FQ ranker in this work which helps to select words based on their frequency to be used as prompts. In contrast, we focus on words unique to specifically scientific domain by removing all words commonly existing in a general dataset.
% In contrast, we focus on words unique to specifically scientific domain by removing all words commonly existing in a general dataset. 
% Additionally, we present an approach to automatically generate relevant time synchronized slides for an existing large dataset for training in contrast to their manual approach. 
% Other work such as LCB-Net \cite{yu2024lcb} propose a novel long-context biasing network for AVSR to leverage the long context information.

Methods of data augmentation has been proposed to create synthetic data with variations of audio and visual modality for the purpose of enhanced speech recognition~\cite{oneațua2022improving}.
In this work, we augment an existing speech-only dataset and enrich them with visual modality for the purpose of multi-modal ASR. 
% {\color{blue} The strategy of automatic data augmentation enables us to use existing large scale domain-specific speech only datasets and enrich them with visual modality for the purpose of multi-modal ASR. }
\cite{chen2024m}, \cite{wang2024slideavsr} present a multi-modal academic dataset for audio-visual recognition and understanding tasks. Both datasets requires manual annotation, which is both time consuming and expensive, making such an approach to large data collection non-feasible. In contrast, we show that ASR model performances can be improved when trained  through an automatically  augmented dataset. While most of the conference videos available are in English, our data augmentation allows utilization of datasets in other languages addition to english.
% In this work, we perform data augmentation by generating slides to mitigate the lack of relevant data. Leveraging the augmented data we perform ASR, thereby enhancing model performance.

\section{Multi-modal Scientific Presentation Benchmark}
In this section we analyze three baseline models on the ability to transcribe on domain-specific words. The models are  evaluated using an evaluation dataset. We describe the dataset in Section \ref{sec:ben_def} and give details of model performance on the dataset in Section \ref{sec:ana}.

\subsection{Benchmark}\label{sec:ben_def}
For evaluating the model performances we use the   
ACL 60/60 dataset \cite{salesky-etal-2023-evaluating}.
This dataset consists of a development (\textit{dev}) and evaluation (\textit{eval}) data each with audio recordings and manual transcripts of technical presentations from ACL 2022 conference. Both the dev and eval sets consist of five recordings each. Each of these datasets has a duration of approximately one hour. The dataset consists of manually created aligned text and audio segments which we consider for our task.
% The second and third is automatically generated by  Supervised Hybrid Audio Segmentation (SHAS)\cite{34946637} and the voice activity detected (VAD) segments respectively. 

\subsection{Metric}
The traditional Word Error Rate (WER) metric is employed to evaluate the performance of ASR models, assigning equal weight to all words in the transcript. In addition to WER, this study places particular emphasis on the ASR performance for words that are frequently encountered within scientific domains. These words are referred to as domain-specific words, and the term special words is used interchangeably throughout this paper.  In this work, we define a domain specific-word as words that does not occure in the general domain corpus (in most experiements this is the Must-C \cite{di-gangi-etal-2019-must} corpus) 
 We measure the quality of the domain-specific words with respect to the reference and the hypothesis similar to recall and precision. First, we investigate how many domain-specific words in the reference are missed or wrongly transcribed by the model, by aggregating the deletion and the substitution counts, and dividing it by the total occurrences of domain-specific words in the manual transcript. 

% To this end, 
% we go beyond WER and perform domain-specific word-only WER.  For analysing baseline model performance, we only consider the domain-specific words in the manually transcribed sentences and count the cases where these words are either deleted, substituted or inserted in the model transcriptions. 
% We investigate how many domain-specific words in the reference are missed or wrongly transcribed by the model, by aggregating the deletion and the substitution counts, and dividing it by the total occurrences of domain-specific words in the manual transcript. 
In this paper, we calculate a reference-centric WER metric $\text{WER}_{t_{ref}}$. 

%  

% \begin{math}
% \mbox{WER-terms}=\frac{\mbox{deletions } + \mbox{ substitutions}} {\mbox{|domain-specific words|}}
% \end{math}
\begin{math}
\mbox{$\text{WER}_{t_{ref}}$}=\frac{\mbox{|substituted} +\mbox{deleted|}} {\mbox{|recognized }+\mbox{substituted}+\mbox{deleted|}}
\end{math}

Next, we calculate the $\text{WER}_{t_{hyp}}$ to evaluate how many domain-specific words in the model's output are incorrectly transcribed.  
% first, by aggregating all the substituted and the deleted counts and then divide it by the total occurrence of domain-specific words in the model transcript.
% In this paper, we calculate a hypothesis-centric WER referred to as $\text{WER}_{t_{hyp}}$.

\begin{math}
\mbox{$\text{WER}_{t_{hyp}}$}=\frac{\mbox{|substituted} +\mbox{inserted|}} {\mbox{|recognized }+\mbox{substituted}+\mbox{inserted|}}
\end{math}

% rec=140, sub=55, del=5, ins=7
% g: the words that exist in the reference....
% t: the works that exited on reference.....
% ref count=200
% hyp count=202
% wer-term-ref ------ 60/200=0.3
% wer-term-hyp------ 62/202=0.3
% normal wer---- 67/200 = 0.3

% rec=100, sub=55, del=7, ins=45
% g: the words that exist in the reference....
% t: the works that exited on reference.....
% ref count=162
% hyp count=200
% wer-term-ref ------ 62/162= 0.38
% wer-term-hyp------ 100/200= 0.5

% rec=100, sub=55, del=45, ins=7
% g: the words that exist in the reference....
% t: the works that exited on reference.....
% ref count=200
% hyp count=162
% wer-term-ref ------ 100/200= 0.5
% wer-term-hyp------ 62/162= 0.38
% normal wer---- 107/200 =0.5

% rec=140, sub=55, del=5, ins=45
% g: the words that exist in the reference....
% t: the works that exited on reference.....
% ref count=200
% hyp count=240
% wer-term-ref ------ 60/200= 0.3
% wer-term-hyp------ 100/240= 0.42
% normal wer---- 105/200=0.5

% rec=45, sub=150, del=5, ins=7
% g: the words that exist in the reference....
% t: the works that exited on reference.....
% ref count=200
% hyp count=202
% wer-term-ref ------ 155/200= 0.78
% wer-term-hyp------ 157/202= 0.78
% normal wer---- 162/200 = 0.8

\subsection{Baseline}\label{sec:base}
To study the ability of ASR models to transcribe domain-specific words we use the models, Whisper, SALMONN and Phi-4-multimodal. 
\paragraph{Whisper:}
% Whisper is a transformer based encoder decoder model created by OpenAI, mainly to perform the task of automatic speech recognition and translation \cite{radford2023robust}. 
% It is trained on about  680k hours of speech data collected from the internet. Whisper encodes the input speech and generates audio features in its encoder part, which  is  eventually forwarded to the decoder. The decoder takes in the audio features along with positional encoding and produces transcription for the input audio. Additionally, Whisper also takes help from  its previous transcriptions for generating the current transcription. In this paper, we use Whisper Large V2 model. 

Whisper is a transformer-based encoder-decoder model developed by OpenAI for ASR and translation tasks \cite{radford2023robust}. Trained on approximately 680k hours of web-sourced speech data, it encodes input audio into features, which are then processed by the decoder to generate transcriptions using positional encoding and prior outputs. In this work, we use the Whisper Large V2 model.
\paragraph{SALMONN:}
% The SALMONN model, developed at Tsinghua University and ByteDance \cite{tang2023salmonn}, empowers Large Language Models (LLMs) like Vicuna \cite{chiang2023vicuna} with the ability to directly perceive and understand general audio inputs. This enables them to achieve competitive performance on various speech and audio processing tasks. The model employs a window-level Q-Former \cite{zhang2024vision} module to integrate the outputs from two encoders: Whisper \cite{radford2023robust} for speech and BEATs \cite{chen2022beats} for general audio. These combined outputs, referred to as augmented audio tokens, are then aligned with the LLM's internal representation. For experiments in this paper, we use  SALMONN 13B v1 model.

The SALMONN model, developed by Tsinghua University and ByteDance \cite{tang2023salmonn}, extends LLMs such as Vicuna \cite{chiang2023vicuna} to directly process and understand general audio inputs, enabling strong performance on various speech and audio tasks. It integrates outputs from Whisper \cite{radford2023robust} and BEATs \cite{chen2022beats} encoders using a window-level Q-Former module \cite{zhang2024vision}, producing augmented audio tokens aligned with the LLM’s internal representations. In this work, we use the SALMONN 13B v1 model.

\paragraph{Phi-4-multimodal:}
% Phi-4-multimodal  referred to as Phi in this paper is a 5.6 billion parameter instruction-tuned multi-modal transformer model developed by Microsoft. It is designed to process  text, image, and audio inputs to support scenarios involving (vision + language), (vision + speech), (speech + language) within an unified architecture. The model supports a context length of up to 128,000 tokens. 
% It incorporates 32 transformers with Group Query Attention(GQA)\cite{ainslie2023gqa} which optimizes memory usage for long-context generation.
% For mapping the vision feature to the text embedding dimension a two layer MLP projector is used. The audio features are also mapped to the text features similarly. 
% Phi demonstrates strong performance across multilingual and multi-modal benchmarks, including visual question answering and automatic speech recognition.
Phi-4-multimodal (referred to as Phi) is a 5.6B-parameter, instruction-tuned multi-modal transformer developed by Microsoft. It supports unified processing of text, image, and audio inputs for vision-language, vision-speech, and speech-language tasks, with a context length of up to 128K tokens. The model employs 32 transformer layers with Grouped Query Attention (GQA) \cite{ainslie2023gqa} for efficient long-context handling. Vision and audio features are projected into the text embedding space using two-layer MLPs. Phi achieves strong performance across multilingual and multi-modal benchmarks.

\subsection{Analysis}\label{sec:ana}

\begin{table*}[ht]
  \caption{Statistics of domain-specific words}
  \label{tab:data_stat}
  \centering
  % \resizebox{10cm}{!}{%
    % \resizebox{\linewidth}{!}{%
    \resizebox{0.9\textwidth}{!}{%
  \begin{tabular}{l|ll|ll|ll|ll}
    \toprule
       \textbf{ }    &  \textbf{} &  \textbf{} & \textbf{Whisper}   & \textbf{} & \textbf{SALMONN}   & \textbf{} & \textbf{Phi}   & \textbf{}   \\
    \midrule
     
      \textbf{ Data} & \textbf{Total} &  \textbf{Unique}   & \textbf{Times}   & \textbf{Times not} & \textbf{Times}   & \textbf{Times not}& \textbf{Times}   & \textbf{Times not}   \\ 

         & \textbf{special words}   & \textbf{special words}  &\textbf{recognised}     & \textbf{recognised} &\textbf{recognised}     & \textbf{recognised} &\textbf{recognised}     & \textbf{recognised}  \\
    
        % & \textbf{ words}   & \textbf{ words}        &   & & &\\
    \midrule
               
          \textbf{ACL dev}                &  333  &  130 & 251  & 82 & 204  & 129 &  244 & 89     \\

    \midrule
                       
   \textbf{ACL eval}       &    276 &    115   & 150 & 126 & 116 & 160 &150 & 126  \\

    \bottomrule  
    \end{tabular}}
\end{table*}

\begin{table}[ht]
  \caption{WER, $\text{WER}_{t_{ref}}$ and $\text{WER}_{t_{hyp}}$ for Whisper, SALMONN and Phi.}
  \label{tab:wer}
  \centering
  \resizebox{\linewidth}{!}{%
  % \resizebox{0.5\textheight}{!}{%
  \begin{tabular}{l|lll|lll}
  
    \toprule
    \textbf{Model}      & \textbf{ACL dev} & \textbf{} & \textbf{}    & \textbf{ACL eval}    & \textbf{}    \textbf{}      \\
    \hline
                & \textbf{ WER } & \textbf {$\text{WER}_{t_{ref}}$} & \textbf {$\text{WER}_{t_{hyp}}$} & \textbf {WER} & \textbf {$\text{WER}_{t_{ref}}$} & \textbf {$\text{WER}_{t_{hyp}}$}    \\
                % & \textbf{  } & \textbf {terms-} & \textbf {terms-} & \textbf{  } & \textbf {terms-} & \textbf {terms-}   \\
                % & \textbf{  } & \textbf {ref} & \textbf {hyp} & \textbf{  } & \textbf {ref} & \textbf {hyp}   \\
  \hline
   % Baseline         & 7.70  & 13.4 & 13.06  & 43.75 
   
    Whisper        & 8.81  & \textbf{24.62} & \textbf{20.57} & \textbf{13.45}  & \textbf{45.65}     & \textbf{44.03}      \\ 
    SALMONN           &   17.42 & 38.44 & 37.31& 20.31 &  57.97   & 57.04\\
    Phi            &   \textbf{7.01} & 26.73 & 25.38 & 18.58 &  45.65  & 44.65  \\

    \bottomrule
  \end{tabular}}
\end{table}
% do you mean prompting in inference? 
% how do we talk about special words in prompts before even saying how we extracted the

% {\color{blue}. 
% also compare with Phi4, make the changes with respect to table1 and table2}

We evaluate the models on their ability to transcribe the ACL dataset specifically on domain-specific words.
% We find that for all models, the word error rate (WER-terms) on domain-specific words is significantly higher compared to WER on all words. 
 % Among the models tested, Whisper consistently achieves the best performance across all cases. 
% {\color{blue}Domain-specific words in this work refer to words existing within scientific context.}
% We select such words by removing all common words from the ACL dataset. 
% Common words are obtained from a general purpose dataset~\cite{di-gangi-etal-2019-must} and we filter such words from the ACL transcripts. The remaining words in the transcript are considered domain-specific words for the purpose of this analysis.

Table \ref{tab:data_stat} gives the  statistics on the domain-specific words extracted from the dataset with this approach. The count of total special words in the ACL dev dataset is 333 of which 130 are  unique. Similarly, there are in total 276 special words in  the ACL eval dataset of which 115 are unique.

The results of the model performance on the ACL dataset are summarized in Table \ref{tab:wer}. We find that for all models, the word error rate ($\text{WER}_{t_{ref}}$ and $\text{WER}_{t_{hyp}}$) on domain-specific words is significantly higher compared to WER on all words.
Whisper makes approximately three times more mistakes on ACL dev and eval datasets.
% whisper mostly don't produce any transcription for these terms. 
% For the Phi model, we observe that it produces approximately 3.4 times more errors on the domain-specific words of ACL dev and about 2.5 times more errors on such words of ACL eval in comparison to the overall WER. 
Similar results can be also observed for SALMONN and Phi models which implies that all models consistently make more mistakes while transcribing domain-specific words.
 Additionally, since $\text{WER}_{t_{ref}}$ and $\text{WER}_{t_{hyp}}$ are similar, there appears to be no specific problem with over or under-generating domain-specific words.

% {\color{blue}
% Although Phi generates transcription with higher error rates than Whisper, as evident in the second row of Table \ref{tab:wer}, it outperforms Whisper for ACL dev in terms of WER. This improvement can be attributed to the fact that, unlike Whisper, Phi generates hallucinations, repetitions, and transcribes numbers as words, which matches the reference. 
% One other difference we observe is when transcribing numbers, phi tends to generate them in words which matches ACL datset reference unlike Whisper which generates numbers in the form of digits. 
% However, Whisper outperforms Phi for ACL eval, because this dataset presents greater transcription challenges for all models as reflected in the WER scores of Table \ref{tab:wer}.
% Given that Whisper is the SOTA ASR model, it is able to transcribe more accurately for ACL eval than PHI across all words.
% }

% The  ACL eval set is challenging for the model compared to the ACL dev set, 
% since both the models gives an increased WER for ACL eval.  The Whisper model works best for all the cases. 
%  Table \ref{tab:wer} shows the result of baseline model performances on the ACL datasets.

We also present the number of times the models are able to recognize the special words.
Columns \textit{Times recognized} and \textit{Times not recognized} of Table \ref{tab:data_stat} show the details of how many of the domain-specific words are recognized and not recognized by  Whisper, Phi and SALMONN models respectively.
We find that Whisper identifies the highest number of domain-specific words on both the ACL dev and ACL eval sets compared to all other models. Notably, Phi matches Whisper's performance in recognizing domain-specific words on the ACL eval set.
Whereas the overall results demonstrate that the domain-specific words pose a difficult challenge for state-of-the-art ASR systems. This motivates the integration of additional context like presentation slides.
The following section describes our approach of additional context extraction and integration to models.
% {\color{red} To summarize, all baseline models make up to three times more mistakes while transcribing domain-specific words.}

\section{Multi-modal  Context Extraction and Integration}\label{sec:multi}
Our analysis on Section~\ref{sec:ana} shows that the current  ASR models make up to three times more mistakes while transcribing domain-specific words. 
% To this end, we propose a multi-modal context extraction and integration system. 
% We build our system on top of an existing ASR model and enrich it through multi-modal information. 

% {\color{blue}We employ both a cascaded approach and an end-to-end approach to incorporate additional information into the model. In the cascaded approach, we extract domain-specific contextual information from the generated slides and subsequently integrate into the model. For the end-to-end approach, visual information is directly incorporated by embedding the image alongside the existing input modalities.} Figure~\ref{fig:overview_diagram} provides an overview of both approaches.

% % For the cascaded approach of context integration we follow a three step process. The first step is to generate images similar to presentation slides. In the second step, we obtain text from the images and finally, augment ASR model with the extracted multi-modal information. 

% For the end-to-end approach we follow a two step process. The first step is to generate images similar to presentation slides and the next step is to integrate the images with the input to the model.
% % The {\color{blue}The images generated in the first step are also utilized in our end-to-end integration approach.}
 
% The following section provides the details on our approach to text extraction and integration into models. 

  Based on this analysis, we propose a multi-modal context extraction and integration system. We build our system on top of an existing ASR model and enrich it through multi-modal information. We propose both a cascaded approach and an end-to-end approach to incorporate additional information into the model. In both cases, we focus on ASR systems based on multi-modal foundation models to allow an easy integration of additional context. Figure~\ref{fig:overview_diagram} provides an overview of both approaches.

In the cascaded approach, we represent the important domain-specific terms explicitly as words and provide these words to the ASR system.  In a first step, we obtain text from extracted images. In a second step, we apply additional filtering on these words. Finally, these words are presented as context to the ASR system.

One disadvantage of this approach is that only the text from the slide is represented and that we can be harmed by cascading errors. Therefore, we also investigate the direct integration of the image in an end-to-end fashion. In this case, the image is provided directly as additional context to the multimodal ASR system.
 
The following section provides the details on our approach to text extraction from images and integration into models.

\subsection{Image Frame Extraction}\label{sec:img_frame}
To obtain the relevant context, we begin with the corresponding video recordings of the scientific talks of the  ACL dataset and extract aligned image frames
 % Given video recording of a presentation with slides and the audio of the presentation, our first component is extraction of image frames from the video. 
 (denoted by 1 in Figure \ref{fig:overview_diagram}).
 % In general, video recordings of presentations are not accompanied by their respective slides. 
 % As a result, we extract the image frames from the recorded video presentation.
 Since presentation video recordings are not usually accompanied by their respective slides, we extract frames directly from the recordings.
  Our audio segments are less than 30 seconds, therefore we assume that while demonstrating the content of a particular segment, the presenter uses only one single slide.
 
 For each of the audio files, we use the available audio segments, with their durations and offset timestamps relative to the full recording. This information is used to align segments with the original video and extract a single frame from the midpoint of each video segment.  
 % We derive the information of the segment duration, which is the length of each segment and an offset timestamp indicating its starting timestamp with respect to the full audio file. 
 % Using these information we then map the audio segments to the original video recording to obtain the respective video segments. From each such video segments, we extract one image frame corresponding to the timestamp in the middle of the segment duration. 
 % We use these image frames in the next steps to generate prompts for the pre-trained models.
 The images are then directly integrated into the end-to-end models or processed to extract the specific vocabulary for the cascaded approach.

% \begin{figure}[t]
%   \centering
%   \includegraphics[scale=0.45]{images/llavaacl_1.pdf}
%   \caption{Text generation with LLaVA-NeXT. The model is provide with a instruction prompt along with an image. LLaVA-NeXT generates text based on the provided inputs.}
%   \label{fig:LLaVA}
% \end{figure}

\subsection{Text Extraction}\label{sec:text}
 In the second component, (denoted by 2 in Figure \ref{fig:overview_diagram}) we perform text extraction on the obtained frames from the previous step (Section~\ref{sec:img_frame}). 
 % We follow two methods to perform this task.
 % For the first method, we consider optical character recognition (OCR) on the image frames and extract all possible texts. 
 % OCR is a strategy to convert texts in images into a machine-readable text format.
% For our second method, we use a SOTA multi-modal LLM and investigate its ability to generate relevant text given image frames as input.
% For our case we use, Large Language and Vision Assistant model (LLaVA) \cite{liu2024visual}, trained on language-image instruction-following data to perform tasks that require visual and language understanding. 
To perform this task, we follow   two methods.
% LLaVA-NeXT and a python OCR library. 
 
% For our case we use, Large Language and Vision Assistant model (Llava) \cite{liu2024visual}, trained on language-image instruction-following data to perform tasks that require visual and language understanding. We provide Llava a pair of the previously extracted image frames and a suitable prompt as input, in order to generate information from each of the frames. 
For the first method, we use LLaVa-NeXT \cite{liu2024llavanext} (referred to as Llava in rest of this paper), due to its ability of better visual reasoning and optical character recognition (OCR) capability. OCR is a strategy to convert texts in images into a machine-readable text format.
We provide  the model with previously extracted image frames and a suitable prompt as input (explained in Appendix~\ref{sec:appendix}), to generate information for each provided frame. 

For our second method, we consider a traditional OCR python library Pytesseract \footnote{https://pypi.org/project/pytesseract/} on the image frames and extract all possible texts.
% For later want to explore if the pre-trained image representations can be useful to perform the task of ASR. 
% Figure \ref{fig:LLaVA} shows one example input pair for the model and the generated output text from the model given the input pair.

 The both methods results in a large number of extracted texts, which needs to be filtered further (denoted by 3 in Figure \ref{fig:overview_diagram}). 
Vision-language models (VLMs) are susceptible to hallucination when extracting text from images. To mitigate this issue, we apply a frequency-based filtering strategy: only words appearing at or above a threshold are retained. Subsequently, as the primary motivation behind this is to obtain only domain-specific words. To this end, we filter the extracted text by 
removing all common words. This is done by discarding all words present in a general presentation dataset \cite{di-gangi-etal-2019-must}, resulting in a collection of only domain-specific words.

\subsection{Context Integration}\label{sec:salmonn}
\begin{figure}[t]
  \centering
  % \vspace{3pt}
  \includegraphics[scale=0.38]{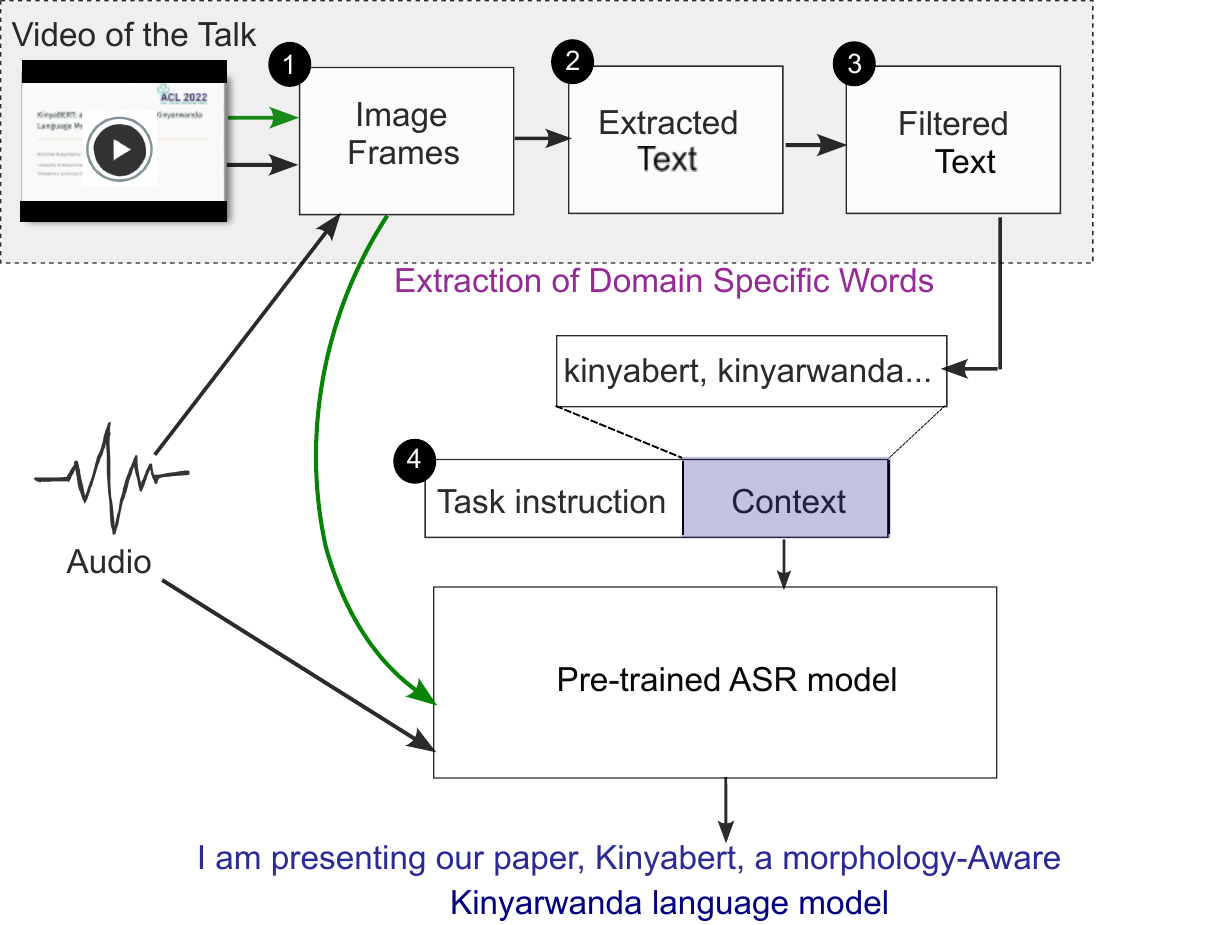}
  \caption{Overview of our two approaches. The green arrows represent the end-to-end approach.}
  \label{fig:overview_diagram}
\end{figure}

The extracted information is then provided to an existing multi-modal ASR model (denoted by 4 in Figure \ref{fig:overview_diagram}). Such ASR systems include an LLM which can be prompted with text to perform the required transcription task. In this work, we focus on improving ASR performance by integrating the context as part of such prompts. 

In particular, we use the additional information to enrich the input to SALMONN and Phi model. By default, there exists  text prompts used in these models that provides instruction (explained in Appendix~\ref{sec:appendix})
% to the integrated LLM Vicuna~\cite{chiang2023vicuna} 
about the task to be performed. We modify the default text prompt with the information extracted from the previous step (Section~\ref{sec:text}).

% \subsection{Image Encoding}\label{sec:image_embedd}

% Encoding an image generally involves converting raw pixel data into a high dimensional feature representation that captures essential visual information. For this study, we use an image encoder capable of integrating with speech and text, enabling flexible fusion of multiple modalities. Additionally, it is important that the encoder is effective for images with rich textual content.

% We perform an end to end training with additional information from image, using a Qformer to covert the encoded image into text tokens.
% For our image input we create a similar Qformer as exists for speech in SALMONN.
% Instead of using a window level Qformer as in speech, image is applied at the frame level, and the output sequence from the Qformer is integrated with the text instruction prompt and fed into the LLM with LoRA adaptors to generate the text response \cite{tang2023salmonn}.
  
% Once the special words for each audio segments are available, we perform the fourth component of our approach
%  (denoted by 4 in Figure \ref{fig:overview_diagram}).

% \section{Integration of Multi-modal information}\label{sec:integrate}

\section{Data Augmentation}\label{sec:approach}
ASR systems with integrated LLMs can be prompted in a zero-shot manner. Existing work \cite{wei2021finetuned} has shown that compared to zero-shot, fine-tuning of models can be useful to achieve further improvements. To this end, we first perform a zero-shot prompting and further enhance the capability of the ASR model to generate accurate transcriptions by incorporating and training with additional information. 

% In order to do that models are required to be trained to use the additional information. 
% We perform data augmentation to existing data with additional slide and use it for training.
% We want the ASR models to gain the ability to generate better transcription, considering the information from slides. 
% We also show that it is not necessary to construct or use  domain-specific dataset for the model to learn how to generate accurate transcription. Instead, existing datasets can be effectively adapted and restructured for this purpose. Furthermore we find that, on training a model with the augmented existing dataset, it gains the ability to generate better transcription beyond domain. 
Enhancing ASR using visual modality, a dataset comprising both visual (e.g. images or slides) and speech data is essential. 
% In this work, we aim to improve ASR model performance on scientific terms by incorporating additional information from conference-based slides.
% However, existing datasets \cite{wang2024slidespeech} are often unsuitable. 
To address the lack of required relevant multi-modal domain-specific data, this work  synthesizes a dataset by data augmentation. For our purpose, we augment images to an existing dataset where we generate images that corresponds to presentation slides. This generated image is then added to the dataset lacking inherent similar multi-modal content. This novel strategy of automatically generating and augmenting a visual modality allows us to use any existing speech dataset while also supporting domain-specific training.  

\subsection{Generation of Presentation Slides}\label{sec:gen}
In this approach, we generate presentation slides for existing speech content through a series of steps. First, we segment the speech transcript into smaller textual units, selecting a chunk size of eight sentences.
% after evaluating several random configurations. 
Our choice of chunk size results in approximately 15–20 slides for a 20–30 minutes speech, ensuring an allocation of 60–90 seconds of speech per slide.

Next, we employ LLaMA 3 to generate LaTeX code for these text chunks. We guide LLaMA 3 with a pair of instructions consisting of a high level system prompt and a more task specific prompt to generate latex code based on the text chunks (explained in Appendix~\ref{sec:appendix}). In the final stage, we convert the generated LaTeX code into images. This involves first compiling the LaTeX code into PDFs and subsequently extracting images from the generated PDF files. We adopt a methodology where images are generated from PDFs rather than directly utilizing the PDFs, as such resources are often unavailable in standard datasets. Conversely, presentation videos are typically accessible, which allows us to extract time-aligned slides corresponding to the speech, as described in Section \ref{sec:img_frame}.    
% pdfs are not time aligned and are also not available

% \subsubsection{Multi-modal Context Extraction}\label{sec:extract}

% This section describes how starting from video files, we extract text that may be used for augmenting the models. 
% We explain this is Section \ref{sec:img_frame}. In a second step, we investigate different strategies to integrate this information into the pre-trained models.
% \supriti{Need to rephrase the following line}

% of data with relevant information
\subsection{Text Extraction }\label{sec:ex}
After obtaining the images from the generated slides, we follow the approach of text extraction by Llava and Pytesseract
as described in Section~\ref{sec:text}. 
Since the target dataset for information augmentation is a general purpose dataset, 
we apply a separate filtration strategy on the extracted text, differing from the one used for the ACL datasets.
% To this end, we first collect all text corresponding to the talks in the dataset.  Next, for each talk, we only keep the text relevant to that particular talk and filter out the remaining text. 
For each talk, we retain only the words that are relevant to the talk by discarding words that also appear in all other talks within the dataset.
We consider such words that are unique to each talk to be the domain-specific words for that particular talk.

\section{Experimental Setup and Results}\label{sec:experiment}
% \supriti{All sections are not mentioned.}
% At first, we perform a zero-shot finetuning on the ASR model.
% Next, along with the existing dataset, we provide additional information  from the generated slides and use these to train the model.

This section provides details on our experimental setup in Section \ref{sec:Experiment_setup} and information about the dataset used for training is included in Section \ref{sec:data}.
% In Section \ref{sec:metric} we specify the evaluation strategies 

\subsection{Experimental Setup}\label{sec:Experiment_setup}
% Since our interest lies in improving the transcription quality of state state-of-the-art models,  
We adopt two models,  SALMONN 13B v1, and Phi-4-multimodal to perform our experiments. 
 For extracting text from the images with LLaVA-NeXT, we use llava-v1.6-mistral-7b model which uses CLIP-ViT-L-336px \cite{radford2021learning} as image encoder and LLaMa \cite{touvron2023llama} for language understanding.  We provide the model with an image as well as a suitable prompt to generate the text from the image.
 
 For generation of slides  we use LLaMa 3 \cite{dubey2024llama} to create latex code and use the python library \textit{subprocess}  to execute the shell commands \textit{pdflatex} and \textit{pdftoppm}  respectively to generate latex code to PDF and image.
 
 % Finally, we also test Whisper on an approach with the best possible outcome, and we refer it as Whisper + Ref prompts for rest of the paper. For this approach, we select all domain-specific words from the manual transcript of the recordings, which is then used as prompt for the respective audio segments.
\subsection{Dataset}\label{sec:data}
For training the ASR model, we use  MuST-C (Multilingual Speech Translation Corpus)  \cite{di-gangi-etal-2019-must} which is primarily  designed as a speech translation dataset. The dataset consists of around 400 hours of audio recordings from English TED Talks speech, transcription and translated transcripts in multiple languages, which are applicable to train model for speech recognition and speech translation tasks. 

Since MuST-C does not contain any visual modality, we augment it with the generated images as described in Section~\ref{sec:approach}. Based on the text extraction and filtration approach described in Section~\ref{sec:ex}, we obtain 16,830 domain-specific words for 2551 talks present in the dataset.

% {\color{blue}Include slidespeech if we put in the paper}

% We want the ASR models to gain the ability to generate better transcription, considering the information from slides. 
% In this work, we show that it is not necessary to construct or use dedicated domain-specific dataset for the model to perform such a task. Instead, existing datasets can be effectively adapted and restructured to serve the intended purpose. Furthermore we find that, on training a model with the augmented existing dataset, it gains the ability to generate better transcription beyond domain

% \begin{figure}[t]
%   \centering
%   \includegraphics[scale = 0.45]{images/dev_seg_new.pdf}
%   \caption{Occurrences of special words recognized by the baseline models and by Whisper and SALMONN with ACL dev segment prompts.}
%   \label{fig:model_recognized_dev}
% \end{figure}
% \begin{figure}[t]
%   \centering
%   \includegraphics[scale = 0.45]{images/eval_seg_new.pdf}
%   \caption{Occurrences of special words recognized by the pretrained model,  SALMONN with prompts, prompt tuned model and the model with end to end training on ACL dataset respectively. }
%   Occurrences of special words recognized by
% the baseline models and by Whisper and SALMONN
% with ACL dev segment prompts.
  \label{fig:model_recognized_eval}
% \end{figure}
  
\begin{table*}[t]
  \caption{Statistics of domain-specific words extracted using Llava, Phi models as well as OCR library Pytesseract} and counts of special words recognized and not-recognized by SALMONN, Phi and Phi+image (Phi trained to perform ASR with image).
  % and Phi models and for Phi trained to perform ASR with image (Phi+image) model. }
  \label{tab:data_stat_all}
  \centering
   % \resizebox[height = 0.50, width= 0.50]{!}{%
  \resizebox{0.9\textwidth}{!}{%
  % \scalebox{0.9}
  \begin{tabular}{l|l|ll|ll|ll|ll}
    \toprule
       \textbf{ } &\textbf{ } & \textbf{}   &  \textbf{} & \textbf{SALMONN}   & \textbf{}& \textbf{Phi}   & \textbf{}  & \textbf{Phi + image}   & \textbf{}   \\ 
    \midrule
     
      \textbf{ Dataset} &  \textbf{Text } & \textbf{Unique}   &  \textbf{Common} & \textbf{Times}   & \textbf{Times not} & \textbf{Times}   & \textbf{Times not} & \textbf{Times}   & \textbf{Times not}   \\ 

     &\textbf{source}    & \textbf{special words}    &  \textbf{with ref} &\textbf{recognised}     & \textbf{recognised}&\textbf{recognised}     & \textbf{recognised}  &\textbf{recognised}     & \textbf{recognised}   \\
    
       % & & \textbf{ words}      &   \textbf{reference }  & &   & & \\
    \midrule
               
                  \textbf{ACL}  &ref        &  130 & - & 204  & 129 & 244  & 89  & 278  & 55  \\
                 \textbf{dev}      &Phi        &  321 & 86 & 164  & 96& 193  & 67  & 218  & 42   \\

                   &Llava            &  367& 81   & 173 &   96& 204 &   65 &231 &   38   \\
                   & \textbf{Pytesseract}           &  475 & 74    & 165  &  77  & 180  &   62 & 205  &  37   \\
    \midrule
                       
   \textbf{ACL}   &    ref    &    115 &  - &    116 & 160&    150 & 126  &    179 & 97 \\ 
    \textbf{eval}                    &Phi         &  645 & 60   &  77 &   108 &  103 &   82 &  124 &   61   \\
    &Llava           &  669 & 60    &  73 &   107 &  95 &   85 &  125 &   55  \\
    &\textbf{Pytesseract}          &  866 & 42    & 56  &   59 & 63  & 52   & 76  &   39  \\
   
    \bottomrule  \end{tabular}}
\end{table*}

\section{Results}\label{sec:result}
In this section we first analyse the quality of the text extracted using Llava,  Phi and Pytesseract in Section \ref{sec:quality_of_llava_extracted_text}. Next, we describe the zero-shot performances of the model on the extracted text presented in Section \ref{sec:zero_shot_performance_of_ASR_model} and finally we compare the zero-shot performance of the model to a model fine-tuned using the additional information elucidated in Section \ref{sec:fine_tuning_performance_using_augmented_data}.

\subsection{Quality of the extracted text} \label{sec:quality_of_llava_extracted_text}
We perform an analysis to check the quality of the extracted text from the images using Pytesseract, Llava and Phi models. This assessment is essential, as the extracted text is intended to support the model's transcription of domain-specific terms.  For this, we compare the special words that are present in the reference text to the extracted text. Table~\ref{tab:data_stat_all} summarizes this result. We find that both the Llava and the Phi model produces a large number of unique special words of which 62\% and 66\% are common to the special words present in the reference of ACL dev  and 52\% is common to the special words in reference of ACL eval dataset. 
% This represents an overlap of 62\% and 66\% with the reference text for SALMONN and Phi models on ACL dev  whereas  52\% overlap between the both both models and the reference words.} The reason for extracting large number of unique special words is because an image that corresponds a slide usually contains additional text that is not uttered by the speaker and as a result not present in the transcript. 

We also measure the performance of the ASR models on the Llava and Phi and Pytesseract extracted words. The considered models for our qualitative analysis are SALMONN, Phi and Phi+image (Phi trained to perform ASR with image) shown as separate columns in Table~\ref{tab:data_stat_all}. As an example, consider the ACL dev dataset where Phi extracted text contains 86 unique special words common to the reference. These 86 words are present in total 260 times in the dataset. The results presented for each ASR models show the number of times out of 260, it has been recognized and not recognized. Consider the results for SALMONN which is able to recognize the Phi extracted special words 164 times but fails for 96 times. Similar ASR model performance results are shown in Table~\ref{tab:data_stat_all} for the Llava extracted text,  traditional OCR pytesseract extracted text and the reference.
% We also measure how the ASR models perform on the extracted text from both models in comparison to the reference. 

% \textbf{\color{red}restructure from here}
% approx same amount  of special words  are extracted by both methods.

% Three models used among which Phi+image 

% models recognizes more words when  llava words for ACL dev although more words are common with ref for phi  and phi for Acl eval. 

% We find that for ACL dev, 81 unique special words overlap with the reference text and is present in total 269 times of which 173 times is recognized by SALMONN while 96 is not recognized. Similarly, for ACL eval, the total number of the unique special words is 180, of which 73 times it is recognized by the ASR model while 107 times it is not. {\color{blue}Out of the 269 total occurrences of special words in ACL dev extracted by Llava, the Phi-4-multimodal model correctly identifies 150, while 126 remain unrecognized. In the case of ACL eval, out of 180 special word instances, the model recognizes 95 and fails to recognize 85.}

\begin{table}[ht]
  \caption{
  WER, $\text{WER}_{t_{ref}}$ and $\text{WER}_{t_{hyp}}$ scores using context words from Llava, Phi, Pytesserect and reference for SALMONN and Phi zero-shot approaches. 
  % WER, $\text{WER}_{t_{ref}}$ and $\text{WER}_{t_{hyp}}$ scores of different setup using SALMONN and Phi, the pre-trained model, zero-shot with Llava prompts, zero-shot with Phi prompts and zero-shot with domain-specific words from the reference transcript.
  }
 
  \label{tab:zero_shot}
  \centering
  \resizebox{\linewidth}{!}{%
  % \scalebox{0.9}
  % \begin{adjustbox}{width=\columnwidth,center}
  \begin{tabular}{l|lll|lll}
  
    \toprule
    \textbf{Model}      & \textbf{ACL dev} & \textbf{} & \textbf{}   & \textbf{ACL eval} & \textbf{}   & \textbf{}          \\
    \hline
    \textbf{ }  & \textbf{ WER } & \textbf {$\text{WER}_{t_{ref}}$} & \textbf {$\text{WER}_{t_{hyp}}$} & \textbf {WER} & \textbf {$\text{WER}_{t_{ref}}$} & \textbf {$\text{WER}_{t_{hyp}}$}    \\
    % & \textbf{  } & \textbf {terms-} & \textbf {terms-} & \textbf {} & \textbf {terms-}& \textbf {terms-}\\
    % & \textbf{  } & \textbf {ref} & \textbf {hyp} & \textbf {} & \textbf {ref}& \textbf {hyp}\\
  \hline
   
     SALMONN        & 17.42  & 38.44 & 37.31 & 20.31  & 57.97 & 57.04\\ 
   
     \hline
    
      + LlaVA prompts         & \textbf{10.31}  & 28.62 & 28.09 & \textbf{16.54}  &  \textbf{48.33}  &    \textbf{47.75}   \\
      + Phi prompts         & 15.36  & \textbf{27.69} &\textbf{ 27.13} & 28.08 &  58.38    &  57.92  \\
     
     + Pytesseract prompts    & 15.93 & 31.82 & 30.96 & 28.96   &  51.30   &   51.30      \\

         + Ref prompts           & 10.93 & 17.12 & 20.66 & 14.09  &  35.87          &      34.93         \\

          \bottomrule
         Phi         & 7.01  & 26.73 & 25.38 & 18.58  & 45.65 & 44.03\\ 
         
\hline
    
      + LlaVA prompts         & \textbf{6.95}  & 21.18 & 20.0& 18.29  &  38.9     &  38.20  \\
      + Phi prompts         & 7.05  & \textbf{20.38} & \textbf{19.46} & \textbf{15.62}  &  38.38   &   37.36  \\
     
     + Pytesseract prompts     & 9.06 & 22.31 & 20.34 & 20.23  & \textbf{32.17}   &   \textbf{32.17}      \\

     + Ref prompts           & 7.01 & 18.02 & 14.95 & 12.30  &  37.68      &  36.06                 \\
    \bottomrule
  \end{tabular}}
\end{table}

\subsection{Zero-shot performance of the ASR model on the extracted data} \label{sec:zero_shot_performance_of_ASR_model}
We evaluate the zero-shot performance of SALMONN and Phi models providing the extracted domain-specific words as prompts and compare it to the model without any additional prompts. 

Table \ref{tab:zero_shot} shows the results of these experiments. It includes our experiments with two models in five configurations. The first configuration referred to as base configuration is the models without any additional prompts shown in first and sixth row of the table. The remaining four configurations considers model with additional context using Llava, Phi, Pytesseract and from the reference text. We conduct experiment using the special words from reference to show the model performance in the best possible configuration.

We find that the model configurations containing additional context  outperforms the base configuration. 
For the SALMONN model the configuration containing Llava context outperforms the base configuration by 26\% and 25\% on ACL dev and 17\% and 16\% on ACL eval on $\text{WER}_{t_{ref}}$ and $\text{WER}_{t_{hyp}}$ respectively. For the Phi model the configuration with additional context extracted from Phi achieves the best results. It outperforms the base configuration by  24\% and 23 \% on ACL dev and by 16\% and 15\% on ACL eval on $\text{WER}_{t_{ref}}$ and $\text{WER}_{t_{hyp}}$ respectively.  

% {\color{blue} The SALMONN model the configuration containing Pytesseract context \ldots}

The SALMONN configuration with Phi context as well as the  Pytesseract context perform poorly on ACL eval in comparison to the base configuration. In contrast, we find consistent improvements over the base configuration for the models when special words obtained from Llava are considered. As a result, for further experiments presented in the paper, we only consider special words from Llava.
 % Both WER and the WER-terms for SALMONN decrease by around 30\% to 40\% for ACL dev set when prompted with Llava extracted text and around 37\% and 55\% when prompted with reference text as shown in last row of  Table \ref{tab:zero_shot}. Similar improvements on WER and WER-terms are also observed for the ACL eval dataset. For this experiment, we use the special words obtained from the reference as prompts to show the model performance when prompted in the best possible setting. 

% We observe that by using special words as prompts, the performance of the models improve significantly. 

\begin{table}[ht]
  \caption{WER, $\text{WER}_{t_{ref}}$ and $\text{WER}_{t_{hyp}}$ scores of different setup using SALMONN and Phi.
  % the pre-trained model, zero-shot with Llava prompts, Fine-tuned with no additional context and the Fine-tuned model with additional information from Llava. The row Fine-tuned with ref shows the best possible setup where the model is fine-tuned using domain-specific words from the reference transcript.}
  % WER refers to the word error rate on all words and WER-terms, refers to error rate on domain-specific words.
  }
  \label{tab:salmonn_wer_scores}
  \centering
  \resizebox{\linewidth}{!}{%
  \begin{tabular}{l|lll|lll}
  
    \toprule
    \textbf{Model}      & \textbf{ACL dev} & \textbf{} & \textbf{}    & \textbf{ACL eval}    & \textbf{} & \textbf{}        \\
    \hline
     & \textbf{ WER } & \textbf {$\text{WER}_{t_{ref}}$}& \textbf {$\text{WER}_{t_{hyp}}$} & \textbf {WER} & \textbf {$\text{WER}_{t_{ref}}$}   & \textbf {$\text{WER}_{t_{hyp}}$}  \\
                    % &  & \textbf {terms-}& \textbf {terms-} &  & \textbf {terms-}& \textbf {terms-}\\
                    %  &  & \textbf {ref}& \textbf {hyp} &  & \textbf {ref}& \textbf {thyp}\\
  \hline
   \textbf{SALMONN}   &  &     & &                 &       &    \\
   
  \hline
  
        Zero-shot   & 17.42 &     38.44 & 37.31&    20.31              &     57.97  &  57.04  \\

         Zero-shot Llava      &  10.31 &    28.62 & 28.09 &   16.54            &   \textbf{48.33}    &  47.75 \\

         Zero-shot Pytesseract           &  15.93  & 31.82 & 30.96 & 28.96  &  51.30   &  51.30       \\
        
        Fine-tuned     & 10.9 &      30.33  & 25.48 & 15.74          &   51.45         & 50.0   \\

        Fine-tune with Llava &   10.24 & \textbf{19.33} & \textbf{17.80}&    \textbf{14.85}          &    48.89    &  \textbf{46.82}  \\

        Fine-tune with Pytesseract &   \textbf{10.07} & 21.49 & 19.83 &            16.75 &    48.70     & 46.85  \\
        
        Fine-tuned with ref  & 9.67 &     10.51  & 8.31 &  14.63             &     29.35    &  26.13\\
   
     \hline
      \textbf{Phi}   &  &     & &                 &       &    \\
  
  \hline
  
        Zero-shot   & 7.01 &     26.73 & 25.38 &    18.58              &     45.65   &  44.03 \\

         Zero-shot Llava      &  \textbf{6.95} &    21.18 & 20.0 &   18.29            &  38.9     & 38.20 \\

         Zero-shot Pytesseract           &  9.06 & 22.31 & 20.34 & 20.23  &  32.17   &   32.17       \\
        
        Fine-tuned     & 9.03 &      22.30 & 20.83 &   13.99          &   40.22     &   39.33     \\

        Fine-tune with Llava &   8.81 & 17.41 &15.85 &    13.66         &    35.0     &  33.52 \\
         
        Fine-tune with Pytesseract &   8.87 & 16.94 & 15.55 &            13.33 &    31.53     &  30.91 \\
       
        Fine-tune with image &   8.70 & \textbf{14.13} & \textbf{13.48}&    \textbf{12.23}          &    \textbf{30.56}     &  \textbf{30.17} \\
        Fine-tuned with ref  & 6.73 &     16.22  & 14.68 &  18.70         &     41.30   & 40.22  \\
    \bottomrule
  \end{tabular}}
\end{table}

\begin{figure*}[t]
  \centering
  \includegraphics[scale = 0.52]{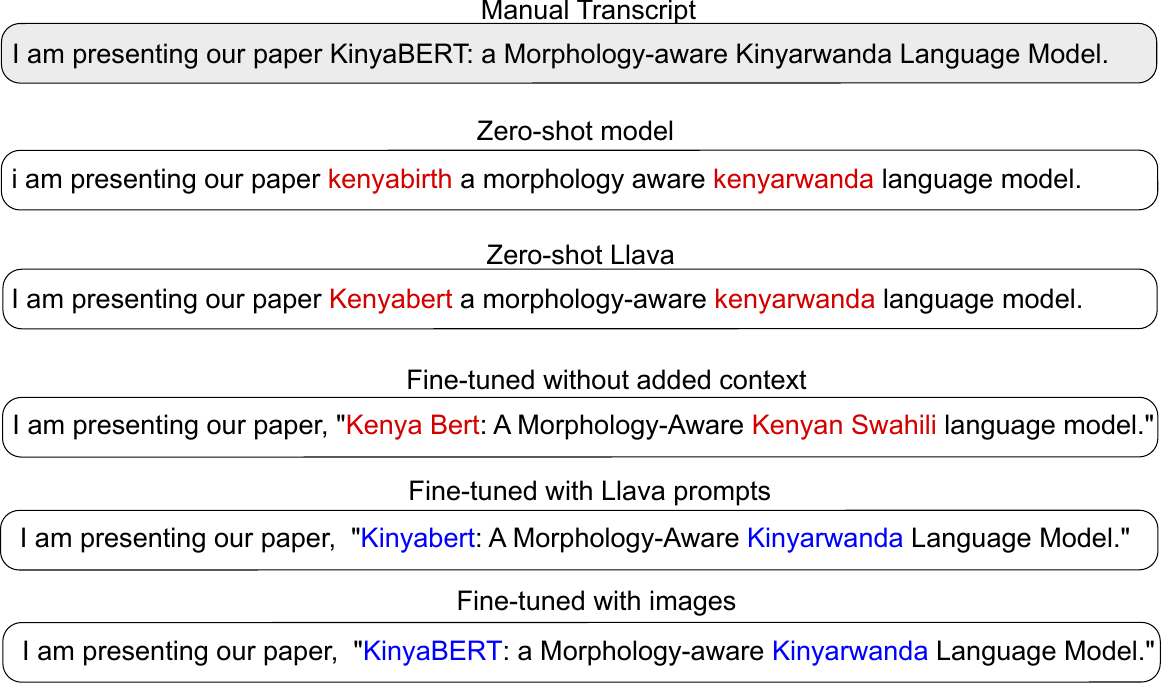}
  \caption{Example of transcriptions generated by different models with respect to the manual transcript. The figure shows that the best possible transcript is generated while fine-tuning the ASR model with llava prompts and image. }
  \label{fig:model_ability_enhancement}
\end{figure*}

\subsection{Fine-tuning performance using augmented data}\label{sec:fine_tuning_performance_using_augmented_data}
For this experiment, our goal is to check if the performance of the ASR models can be improved further by fine-tuning compared to zero-shot performance. To this end, we fine-tune SALMONN and Phi using the augmented dataset obtained in Section~\ref{sec:approach} and compare it to  additional setups described below. 
Table~\ref{tab:salmonn_wer_scores} summarizes the results of our experiment.

The upper part of the table illustrates the SALMONN specific setups and their corresponding results while the lower part contains the Phi specific details. 
% We fine-tune the models using the domain-specific words from reference (\textit{Fine-tuned with ref}) which represents our best possible setup. 
% We include the model's performance once using augmented dataset with Llava extracted context (\textit{Fine-tuned with Llava}) and using no context (\textit{Fine-tuned}). 
% The table shows that the model improves on its performance when fine-tuned with Llava context words over the model trained with no context words. 
% Furthermore, we also show that both the model's performance improves with all fine-tuning methods with respect to zero shot experiments Table~\ref{tab:salmonn_wer_scores}.
% WER and WER-terms scores of different setup using SALMONN and Phi, the pre-trained model, zero-shot with Llava prompts, Fine-tuned with no additional context and the Fine-tuned model with additional information from Llava. The row Fine-tuned with ref shows the best possible setup where the model is fine-tuned using domain-specific words from the reference transcript.
The following provides details on the setups of our experiment that corresponds to Table~\ref{tab:salmonn_wer_scores}.
\begin{description}
    \item[] \textit{Zero-shot Llava:} The model with additional context using Llava  (Section~\ref{sec:zero_shot_performance_of_ASR_model}).
    \item[] \textit{Zero-shot Pytesseract:} The model with additional context using the OCR library Pytesseract  (Section~\ref{sec:zero_shot_performance_of_ASR_model}).
    \item[] \textit{Fine-tuned:} The model fine-tuned without any additional context using the configurations used by the model authors i.e., no changes are made to the task description. (Section~\ref{sec:zero_shot_performance_of_ASR_model}).
    \item[] \textit{Fine-tuned with Llava:} The model fine-tuned with additional context words from Llava (default setup).
    \item[] \textit{Fine-tuned with Pytesseract:} The model fine-tuned with additional context words using Pytesseract .
    \item[] \textit{Fine-tuned with image:} The model (only done for Phi since it accepts image as input) fine-tuned with image instead of additional text as context.
    \item[] \textit{Fine-tuned with ref:} The model fine-tuned with context obtained as special words from transcripts (best possible setup). 
\end{description}

For the setups mentioned above that uses additional context words, we modify the model's task description with additional special words and change the instruction to consider the special words while transcribing (explained in Appendix~\ref{sec:appendix}). Additionally, we make sure that during extraction of special words as outlined in Section~\ref{sec:approach}, there exists no overlap between special words from training and evaluation datasets.  

% The first row of  the table (\textit{denoted by fine-tuned with ref}) shows the best possible setup when the models are fine-tuned with domain-specific words extracted from the reference transcript. The second and third setup that corresponding to the second and third row of the tables are described in Section~\ref{sec:zero_shot_performance_of_ASR_model}. 

% The fourth setup shown as \textit{fine-tuned}, records the performance of the models when fine-tuned on the MUSTC dataset without any augmentation. The last row of the table \ref{tab:salmonn_wer_scores} and second to last row of table \ref{tab:phi_wer_scores} shows our default setup of fine-tuning the model using the augmented dataset with Llava prompts. The last row of the table \ref{tab:phi_wer_scores} presents the results of Phi model when it is fine-tuned with images directly, rather than using context derived from those images. 

% \paragraph{Fine-tuning performance} 
As illustrated in Table~\ref{tab:salmonn_wer_scores}, both SALMONN and Phi models improve its overall performance when fine-tuned with Llava context words over fine-tuned with no context words. For the SALMONN setups, fine-tuning with Llava words achieves the best possible scores across both the datasets. We observe, similar results for the Phi setups with additional context words. 
We conduct additional experiments with Phi using image instead of extracted words as addition context. We find this to be our best possible overall setup for Phi, even outperforming the setup containing context words from reference. This improvements can be attributed to the fact that in addition to text in the slides, the included figures, plots and tables also contribute to the model performance.  

We perform a significance test by using matched-pair test for error counts for two hypothesis 1) transcripts from model using only speech 2) transcripts from model using speech and additional context. We find a p-value of less than 0.001 showing the significance of our results.

% Fine-tuning both SALMONN and  Phi for ASR tasks require a task description as an instruction to the integrated Llama model in SALMONN and the Phi-4-Mini in Phi. For the fourth setup, the model is fine-tuned using the configurations used by the model authors i.e., no changes are made to the task description. For our default configuration, we modify this task description with additional special words and change the instruction to consider the special words while transcribing (explained in Appendix~\ref{sec:appendix}). 

% Our results demonstrate that the overall model performance improves on transcribing special words that are not present in the training dataset which shows that the model is not merely remembering the words. 

% -------------------------

Figure \ref{fig:model_ability_enhancement}, shows an example prediction by the Phi model 
with each setup described earlier. Considering both the Zero-shot model and the Fine-tuned model without context words, we find that the models make mistakes on both words \textit{KinyaBERT} and \textit{Kinyarwanda}. The zero-shot with Llava model improves but is unable to transcribe correctly. Whereas the Fine-tuned model with LLava generates the correct transcription  likely due to its acquired ability to incorporate from the additional information. Finally, the model trained with images not only accurately transcribes the content but also preserves the textual formatting as it appears in the presentation slide. As illustrated by the above example, our experiments show encouraging results in improving existing ASR performance either using context words or images. 

To be used for ASR of scientific talks, the approach requires minimal additional effort to setup. An example setup comprises of a system to generate images from slides of a presenter which is directly utilized by the ASR models for improved transcription. 
% considering segment-wise prompt. We see that the GOT-OCR2  approach generates a more plausible word than the baseline,  but the generation is better using the LLaVA approaches. We note that for this particular example, model + Ref prompts also generates the correct transcription.  We refer GOT-OCR2 as OCR  and LLaVa-NeXT as llava for the rest of the paper.
 
% {\color{blue}\section{Analysis }
% To gain deeper insight into how the model utilizes keywords, we specifically count how many special words play a role in correcting the predictions. }

\section{Conclusion and Future work}
% Modern ASR systems do not perform well on transcribing domain-specific words. 
%  This paper, investigates how multi-modal information can be augmented to improve the performances of a SOTA ASR model Whisper. 
%   We extract information from video recordings of slides, and use the extracted information as prompts. We find that Whisper with prompts achieves a lower word error rate on special words.
%   The results in \ref{sec:result} suggest ample opportunity for improvement.
%   As a future work we intend to integrate image representation to the model and further investigate Whisper on ASR performances.
Current ASR systems exhibit challenges in accurately transcribing domain-specific words. This limitation hinders their effectiveness in various applications. We present an analysis of the model performance on transcribing domain-specific words to demonstrate this. 
This paper investigates the potential of augmenting  ASR models with information extracted from slides to improve performance. We explore the use of visual information extracted from video recordings of slides as prompts.
 When trained with additional context, the model develops ability to generate better transcription on domain-specific words. This shows the effectiveness of multi-modal information in enhancing ASR performance.
 
The results presented in Section~\ref{sec:result} highlight the potential for further advancements. We find that integrating image as an additional input improves ASR performances for Phi and as future work propose to investigate on SOTA ASR uni-model performances on such end-to-end approaches.

\section{Acknowledgement}
This work was supported by the European Union’s Horizon Europe Framework Programme under grant agreement No. 101213369, project DVPS (Diversibus Viis Plurima Solvo).

Additional support was provided by KiKIT (Pilot Program for Core-Informatics at KIT) of the Helmholtz Association.

We also acknowledge the use of the HoreKa supercomputer, funded by the Ministry of Science, Research, and the Arts of Baden-Württemberg, and by the Federal Ministry of Education and Research.

% http://doi.org/10.6084/m9.figshare.30158932

% 10.6084/m9.figshare.30158932
% \begin{table}[ht]
%   \caption{\color{blue}Recall scores of different setup using SALMONN and Phi the pre-trained model, zero-shot with Llava prompts, Fine-tuned with no additional context and the Fine-tuned model with additional information from Llava and Phi trained with image.}
%   % WER refers to the word error rate on all words and WER-terms, refers to error rate on domain-specific words.}
%   \label{tab:recall}
%   \centering
%   \resizebox{\linewidth}{!}{%
%   \begin{tabular}{l|l|ll}
  
%     \toprule
%     \textbf{Model}   &   & \textbf{ACL dev}     & \textbf{ACL eval}            \\
%     \hline
%     & & \textbf{ WER }  & \textbf {WER-}     \\
%                     &  \textbf {terms}   & \textbf {terms}\\

%   \hline
  
%         &Zero-shot   & 0.63 &     0.42       \\

%    SALMONN      &Zero-shot Llava      &  0.69 &    0.52   \\
        
%         &Fine-tuned     & 0.70 &      0.46         \\

%         &Fine-tune with Llava &   0.77 & 0.50    \\
%      \hline
%       &Zero-shot   & 0.74 &     0.56       \\

%    Phi      &Zero-shot Llava      &  0.80 &    0.61   \\
        
%         &Fine-tuned     & 0.75 &      0.59         \\

%         &Fine-tune with Llava &   0.80 & 0.63    \\
%         &Fine-tune with image &   0.84 & 0.67    \\

%     \bottomrule
%   \end{tabular}}
% \end{table}

\newpage

\section*{Limitations}\label{sec:limit}
%  important words not in the slides
% Additional error by the vLLM
% Additional computation through vLLM

While our augmented data approach proves effective and results in significant improvements in model performance, it is not without limitations, presenting opportunities for future research. 

In our work we consider slides to extract domain-specific words that can be used as additional information for context integrated ASR. Slides often contains summarized, bullet-pointed information which may lead to omit domain-specific words to some extend which may effect the models ability to recognize them correctly. 
Speakers often elaborate the slides with their own words introducing mismatch between speech and the slide content which also creates similar problem.
Additionally, we use  OCR and vision-language models (VLMs) are susceptible to hallucination when extracting text from images. Although we took measures to mitigate this issue but with the expense of some important domain specific words.
Apart from that, the ASR model in this work integrates a pre-trained LLM.
LLMs are heavily dependent on the quality and diversity of their training data. Although we achieve improved model performance with our augmented data there remains further scope of improvement. When integrating additional information to the LLM, it may fail to effectively combine these sources of information, leading to misaligned predictions  for some cases.
 Incorporating LLMs into the ASR pipeline for context integration introduces substantial computational overhead, which can slow down the processing time. 
 
On the other the LLM might misinterpret the contextual information for the speech and lead to produce incorrect transcription. 

Our experiment involving image integration into the existing ASR model is limited to the Phi-4-multimodal model. Further comprehensive studies are required to draw conclusive insights into model performance under such configuration.

\bibliography{acl_latex}

\section{ Appendix}
\label{sec:appendix}
\appendix
% \subsection{Textual Context Integration }
\paragraph{Textual Context Integration to SALMONN} 
We instruct SALMONN by providing text prompts to Vicuna   that ask questions about the processed audio. The LLM then responds with textual answers based on its understanding. 
The model is trained for various speech related tasks with suitable prompt structure, as follows
\begin{lstlisting}
USER: [Auditory Tokens] Can you transcribe the speech into a written format? \n ASSISTANT:
\end{lstlisting}
Here,\textit{ [Auditory Tokens]}are the output tokens of the window-level QFormer, followed by user prompts in the form of questions with respect to the task performed by the model on the given audio.

Our extracted domain-specific terms from accompanying slides are included in prompts with the following structure
\begin{lstlisting}
USER: [Auditory Tokens] Please can you transcribe the speech referring to the following tokens wherever needed: kinyarwanda, kinyabert, nlp, pre-trained, ...? \n ASSISTANT:
\end{lstlisting}
Here, domain-specific words like \textit{ Kinyarwanda, Kinyabert, NLP, and pre-trained} are included in the user prompt. The overall prompt is designed to emphasize both these special words and the task itself.
% This is an appendix.
\paragraph{ Context Integration to Phi}
Depending on the input required for training Phi-4-multimodal modal, we construct its prompt format.

Format for Speech-Language with special words:

\begin{lstlisting}
user_message = {
    "role": "user",
    "content": "<|audio_1|>\n" + Can you transcribe the given speech referring to the following words wherever needed #### kinyarwanda, kinyabert, nlp, pre-trained, ...?
}
\end{lstlisting}

Format for Speech-image-Language: 
\begin{lstlisting}
user_message = {
    "role": "user",
    "content": "<|image_1|>\n<|audio_1|>\n" + 
    Can you transcribe the given speech?
}
\end{lstlisting}        

\paragraph{Model Instruction for Text Extraction }
To exhibit LLaVa-Next models OCR quality an extract text from slides we provide the model with an image and a suitable text prompt. the structure of the instruction is given as follow:

\begin{lstlisting} 
"[INST] <image>\nUSER: Extract the text from the sides? [/INST]"
\end{lstlisting} 

  the \textit{<image>} tag is replaced with the image input for LLaVa-Next following with the user prompt. The instruction should always start with the  \textit{[INST]} tag and end with  \textit{[/INST]} tag.

\paragraph{Model Instruction for Data Augmentation }
For creating the multi-modal context for data augmentation, we use LLaMa 3 and guide it with a pair of instructions consisting of a high level system prompt and a more task specific prompt to generate latex code based on text chunks.
This consists of a system prompt and a user prompt as follows:
\begin{lstlisting}
{"role": "system", "content": "you are a presenter who wants to inform and inspire"},

{"role": "user", "content": generate one presentation slide with the main points and concepts in latex, from the following text:<chunk>}
\end{lstlisting}    
        The \textit{chunk} in the user prompt is replaced by the parts of talk for which we want to generate the latex code.

\end{document}